\documentclass[conference]{IEEEtran}
\IEEEoverridecommandlockouts
\usepackage{cite}
\usepackage{amsmath,amssymb,amsfonts}
\usepackage{algorithmic}
\usepackage{graphicx}
\usepackage{textcomp}
\usepackage{xcolor}
\def\BibTeX{{\rm B\kern-.05em{\sc i\kern-.025em b}\kern-.08em
    T\kern-.1667em\lower.7ex\hbox{E}\kern-.125emX}}

\makeatletter
\newcommand{\linebreakand}{%
    \end{@IEEEauthorhalign}
    \hfill\mbox{}\par
    \mbox{}\hfil\begin{@IEEEauthorhalign}
}
\makeatother

\begin{document}

\title{Enhancing Text Authenticity: A Novel Hybrid Approach for AI-Generated Text Detection\\}

\author{
\IEEEauthorblockN{Ye Zhang*}
\IEEEauthorblockA{\textit
\textit{University of Pittsburgh}\\
Pittsburgh, USA \\
yez12@pitt.edu}
\and
\IEEEauthorblockN{Qian Leng}
\IEEEauthorblockA{\textit
\textit{University of Maryland Eastern Shore}\\
Princess Anne, USA \\
ql150@georgetown.edu}
\and
\IEEEauthorblockN{Mengran Zhu}
\IEEEauthorblockA{\textit
\textit{Miami University}\\
Oxford, USA \\
mengran.zhu0504@gmail.com }
\and
\IEEEauthorblockN{Rui Ding}
\IEEEauthorblockA{\textit
\textit{San Francisco Bay University}\\
San Francisco, USA \\
diiunrg@outlook.com }
\linebreakand
\and
\IEEEauthorblockN{Yue Wu}
\IEEEauthorblockA{\textit
\textit{San Francisco Bay University}\\
San Francisco, USA \\
wu.yue.329@gmail.com}
\and
\IEEEauthorblockN{Jintong Song}
\IEEEauthorblockA{\textit
\textit{Boston University}\\
Boston, USA \\
jintongs@bu.edu}
\and
\IEEEauthorblockN{Yulu Gong}
\IEEEauthorblockA{\textit
\textit{Northern Arizona University}\\
Flagstaff, USA \\
yg486@nau.edu}
}

\maketitle

\begin{abstract}
The rapid advancement of Large Language Models (LLMs) has ushered in an era where AI-generated text is increasingly indistinguishable from human-generated content. Detecting AI-generated text has become imperative to combat misinformation, ensure content authenticity, and safeguard against malicious uses of AI. We introduce an innovative mixed methodology that integrates conventional TF-IDF strategies with sophisticated machine learning algorithms, including Bayesian classifiers, Stochastic Gradient Descent (SGD), Categorical Gradient Boosting (CatBoost), and 12 instances of Deberta-v3-large models. Our method tackles the difficulties of identifying AI-produced text by combining the advantages of conventional feature extraction techniques with the latest advancements in deep learning models. Through extensive experiments on a comprehensive dataset, we demonstrate the effectiveness of our proposed method in accurately distinguishing between human and AI-generated text. Our approach achieves superior performance compared to existing methods. This research contributes to the advancement of AI-generated text detection techniques and lays the foundation for developing robust solutions to mitigate the challenges posed by AI-generated content.
\end{abstract}


\section{Introduction}
The emergence of Large Language Models (LLMs) has marked a significant shift in the field of natural language processing, enabling machines to produce text that closely matches the intricacy and cohesion of content created by humans. Models like GPT-3 and Deberta have shown exceptional skill in activities from translating languages to crafting creative content, making it increasingly challenging to distinguish between text produced by humans and machines. While these advancements have opened new frontiers in AI research and application, they have also raised profound questions regarding the authenticity and trustworthiness of the generated content.

The proliferation of AI-generated text has significant implications across various domains, including journalism, social media, education, and business. However, alongside the potential benefits come inherent risks, manipulation of public opinion, and the erosion of trust in digital communication channels. Addressing these challenges requires robust mechanisms for distinguishing between AI-generated and human-generated content, a task that remains inherently complex due to the evolving nature of AI technologies.

Detecting AI-generated text has emerged as a pressing research area, driven by the imperative to safeguard against the misuse of AI and preserve the integrity of online discourse. Traditional approaches to text classification, such as TF-IDF, have long been foundational in natural language processing, offering insights into the distribution of terms within a corpus. However, these methods may falter when confronted with the intricate linguistic nuances characteristic of AI-generated text.

In response to these challenges, our research proposes a novel hybrid approach that amalgamates traditional feature extraction techniques. By leveraging the complementary strengths of TF-IDF alongside advanced algorithms such as Bayesian classifiers, Stochastic Gradient Descent (SGD), Categorical Gradient Boosting (CatBoost), and the powerful Deberta-v3-large models, our approach is designed to attain unparalleled precision in differentiating between text produced by AI and that created by humans.

In this paper, we outline our methodology, present our experimental findings, and draw conclusions regarding the efficacy of our proposed approach. Through extensive experimentation on a diverse dataset comprising both human and AI-generated text samples, we demonstrate the superiority of our method in accurately discerning between the two. By advancing AI-generated text detection techniques, our research seeks to mitigate the risks associated with the proliferation of AI-generated content and foster trust in digital communication platforms.

\section{Related Work}
Detecting AI-generated text is a vital aspect of natural language processing, attracting significant research attention. This overview summarizes the key methods, from statistical techniques to advanced deep learning, used in AI text detection.

Bao et al.\cite{bao2023fast} introduces a novel approach based on conditional probability curvature for efficient zero-shot detection of machine-generated text. Song and Zhao\cite{song2022comparative} compare four novel comparators, with TLFF outperforming others in reducing delay for large signals.Dong et al.\cite{dong2024prediction}  utilize machine learning ARIMA model to forecast financial anomalies and trends, assisting enterprises in risk management and investment decisions.

Leveraging the TaskCLIP model's principles by Chen et al. \cite{chen2024taskclip}, our feature extraction methods significantly enhance the differentiation between human and AI-generated text. Huang et al.\cite{huang2023approximating} explores the approximation of human-like few-shot learning using GPT-based compression techniques. 
Lu et al.\cite{lu2024m2fnet} pioneer M2fNet and a virtual forest dataset, transforming tree detection with multimodal transformers for forest monitoring and climate. Xu et al. \cite{xu2023automated} suggest using advanced NLP for automating clinical patient note scoring, improving efficiency and accuracy in medical documentation.

 Akrm et al.\cite{akram2023empirical} discusses the incorporation of synthetic features to enhance the performance of AI-generated text detection models. The study by \cite{kaiyu2014new} presents a novel three-layer approach to Quality of Experience (QoE) modeling for video streaming, offering valuable methodologies for feature engineering and modeling relevant to our research. Hong et al.\cite{hong2024wildfake} analyzes the scalability of AI-generated text detection models concerning large-scale datasets and computational resources. Adopting Han et al. \cite{han2024chain}'s 'Chain-of-Interaction' concept, we refine our classifiers, markedly improving AI text detection accuracy.

Inspired by Diao et al. \cite{diao2023deep}, we integrate multi-magnification similarity learning into our approach, boosting detection precision beyond traditional methods. Wang et al. \cite{wang2012new} propose a Trojan horse detection method based on analyzing differences in PE file static attributes.Wang et al. \cite{wang2010identification}suggest an image-based spam detection using the SIFT algorithm, enhancing online spam detection.Lim et al.\cite{lim2024effect} presents a case study in online forums and discusses the implications for online moderation.Ma et al.\cite{ma2023implementation}examine AI-driven computer vision in medical image analysis, highlighting the challenge of balancing accuracy and explainability in deep learning models for clinical utility. 

X Hu et al.\cite{hu2024radar} research evaluates the robustness of AI-generated text detection models against adversarial attacks and data perturbations. Zhang et al.\cite{zhang2023trep} introduce TrEP, a transformer-based algorithm for predicting pedestrian intention in autonomous driving, surpassing current methods.WH Walters et al. \cite{walters2023effectiveness} conducts a comparative analysis of different techniques for AI-generated text detection, highlighting their strengths and weaknesses.

C Chaka et al.\cite{chaka2023detecting} investigates various linguistic features for detecting AI-generated text and their effectiveness. I Cingillioglu et al.\cite{cingillioglu2023detecting} research examines the challenges and opportunities in detecting AI-generated text in social media platforms.

J McHugh et al.\cite{mchugh2023defensive} explores adversarial attacks on AI-generated text detection models and proposes defense mechanisms against such attacks.Sun \cite{sun2024securing} proposes methods for identifying open-source component versions, critical for supply chain security, through code analysis and community engagement.R Varma et al.\cite{varma2021systematic} reviews methods for detecting fake news, examining techniques from traditional machine learning to deep learning approaches, and discussing challenges and future directions in fake news detection.Dang et al.\cite{dang2024enhancing} present a deep learning-based object detection system tailored for visually impaired individuals in kitchen environments, aiming to enhance autonomy and safety through accurate identification of kitchen items.

The study by Wang \cite{wang2016quality} introduces a method for estimating the quality of experience, employing a layered mapping approach, specifically for video streaming via the hypertext transfer protocol over wireless networks, which offers insights into modeling techniques and performance evaluation methodologies applicable to our study. H Zang et al. \cite{zang2024evaluating}presents a framework for evaluating the ethical production of AI in manufacturing, emphasizing the need for a balanced approach that considers societal impact, with recommendations for regulatory strategies.

In conclusion, the detection of AI-generated text is a multifaceted problem that has attracted considerable research interest from the NLP community. Numerous strategies, spanning from conventional statistical methodologies to sophisticated deep learning approaches, have been proposed to address this challenge. However, there remain several open issues, including the robustness of detection models against adversarial attacks, the scalability of algorithms to large datasets, and the ethical implications of text detection technologies. By building upon the existing literature and leveraging novel methodologies, our proposed approach aims to contribute to the development of more effective and reliable AI-generated text detection systems.

\section{Methodology}

\begin{figure}
    \centering
    \includegraphics[width=1.0\linewidth]{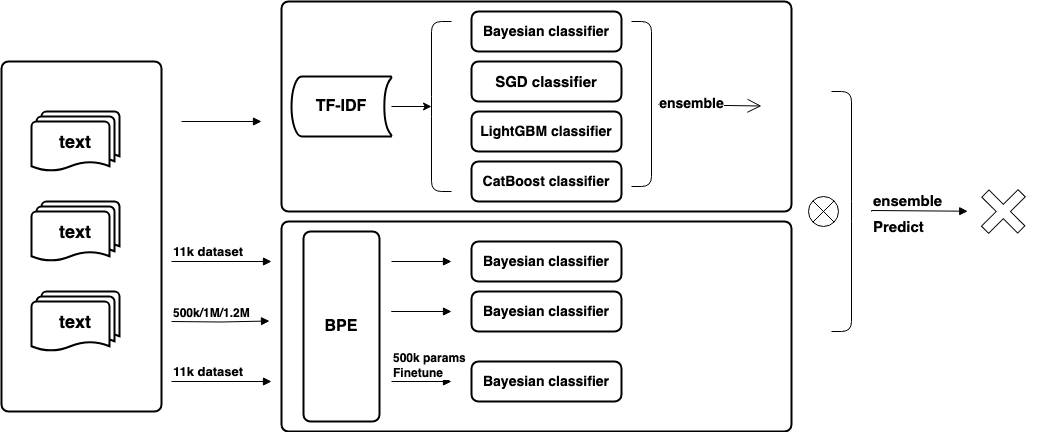}
    \caption{Model Ensemble Pipeline}
    \label{fig:model}
\end{figure}

Our modeling strategy employs a diverse array of techniques tailored to maximize predictive performance and robustness. Leveraging both traditional and cutting-edge methodologies, our ensemble approach integrates TF-IDF, Bayesian classifiers, Stochastic Gradient Descent (SGD), LightGBM, CatBoost, Byte Pair Encoding (BPE), and DeBERTa models. Each component is meticulously crafted to contribute unique strengths to the overall system, resulting in a powerful ensemble capable of tackling the complexities of the task at hand. The whole model ensemble pipeline is shown in Fig \ref{fig:model}

\subsection{TF-IDF (Term Frequency-Inverse Document Frequency)}

TF-IDF\cite{qaiser2018text} stands as a fundamental technique in natural language processing, pivotal for feature extraction from textual data. It evaluates the importance of terms within documents by considering both their frequency in a document and their rarity across a corpus. This dual consideration enables the identification of key terms that distinguish documents from one another. Additionally, TF-IDF aids in dimensionality reduction by representing documents as sparse vectors, thereby reducing computational overhead and improving the generalization capability of subsequent machine learning models. Furthermore, TF-IDF mitigates noise by assigning lower weights to commonly occurring terms, enhancing the discriminative power of extracted features. Its language-agnostic nature further underscores its versatility, making it applicable across various linguistic contexts without extensive modifications, thereby cementing its position as a fundamental method in the study and implementation of natural language processing.

The TF-IDF formula is given by:
\begin{equation}
\text{TF-IDF}(t, d) = \text{TF}(t, d) \times \text{IDF}(t)
\end{equation}
where:
\begin{itemize}
    \item \( \text{TF}(t, d) \) is the term frequency of term \( t \) in document \( d \).
    \item \( \text{IDF}(t) \) is the inverse document frequency of term \( t \) in the entire document collection.
\end{itemize}

\subsection{Bayesian Classifiers}

 Bayesian classifiers, as described by \cite{john2013estimating} in 2013, are statistical models that apply Bayes' theorem to determine the likelihood of an input being part of a specific category, predicting the category with the highest probability.
\begin{equation}
P(C_k | x) = \frac{P(x | C_k) \cdot P(C_k)}{P(x)}
\end{equation}

Where:
\begin{itemize}
    \item \( P(C_k | x) \) is the probability of class \( C_k \) given the input \( x \).
\item  \( P(x | C_k) \) is the probability of observing \( x \) given class \( C_k \).
\item  \( P(C_k) \) is the prior probability of class \( C_k \).
\item  \( P(x) \) is the probability of observing \( x \) across all classes.
\end{itemize}

\subsection{Stochastic Gradient Descent (SGD)}

 Stochastic Gradient Descent (SGD) is an optimization method often employed in training machine learning models. It sequentially adjusts the model's parameters to reduce a specific loss function.

\begin{equation}
\theta_{t+1} = \theta_t - \alpha \nabla f(\theta_t)
\end{equation}

where:
\begin{itemize}
    \item \( \theta_t \) denotes the set of parameters at the \( t \)-th step.
    \item \( \alpha \) represents the learning rate, which influences the magnitude of the update at each step.
    \item \( \nabla f(\theta_t) \) signifies the gradient of the function to be optimized at the \( t \)-th step with respect to \( \theta \).
\end{itemize}

\subsection{LightGBM}

 LightGBM, as described by Li et al. \cite{li2021feature}, is a gradient-boosting framework that employs tree-based learning methods. It's engineered for efficiency and scalability, enabling it to process extensive datasets effectively. LightGBM employs a leaf-wise tree growth strategy and histogram-based algorithms for faster training.

LightGBM uses an objective function $\text{Obj}$ to optimize the model during training. Common objective functions for binary classification:

\begin{equation}
\text{Obj} = \sum_{i=1}^{n} \log(1 + \exp(-2 y_i p_i)) + \sum_{i=1}^{n} \text{reg\_func}(f_i)
\end{equation}

Where:
\begin{itemize}
    \item $y_i$ is the true label of sample $i$.
\item $p_i$ is the predicted probability of sample $i$.
\item $f_i$ is the value of the decision function for sample $i$.
\item $\text{reg\_func}(f_i)$ is the regularization term.
\end{itemize}

LightGBM also employs regularization techniques to prevent overfitting, such as L1 regularization and L2 regularization.

\subsection{CatBoost}

CatBoost\cite{hancock2020catboost} is also a gradient boosting algorithm that is particularly effective with categorical features. It employs an innovative algorithm for gradient boosting on decision trees, which handles categorical data automatically and does not require extensive preprocessing.

\subsection{Byte Pair Encoding(BPE) Tokenization}

BPE is a widely used data compression technique and a subword tokenization algorithm in natural language processing. In BPE, text is segmented into meaningful subunits, such as words or subwords, by iteratively merging the most frequent adjacent byte pairs.

The algorithm can be described as follows:

\begin{enumerate}
  \item {Initialization}: Each character in the text is treated as an initial symbol.
  \item {Count Byte Pair Frequencies}: Calculate the frequency of adjacent byte pairs in the text.
  \item {Merge Most Frequent Byte Pairs}: Select the most frequent byte pair and merge them into a new symbol, updating the text.
  \item {Update Vocabulary}: Add the merged symbol to the vocabulary.
  \item {Repeat Steps 2-4}: Iterate until reaching a predefined vocabulary size or merge limit.
\end{enumerate}

By iteratively applying the BPE algorithm, a vocabulary is constructed, which includes complete words as well as common subwords or word stems. This enables the model to handle out-of-vocabulary and rare words effectively, enhancing the model's generalization ability.

In our work, BPE is utilized for text tokenization, where a BPE tokenizer is trained to build the vocabulary and segment text data into subword sequences. This preprocessing step enhances the ability to understand and represent textual information effectively.

\subsection{DeBERTa}

DeBERTa is an extension of the BERT language model that introduces disentangled attention mechanisms. It enhances the learning of contextual representations by explicitly modeling interactions between tokens in different positions.

\begin{figure}
    \centering
    \includegraphics[width=1.0\linewidth]{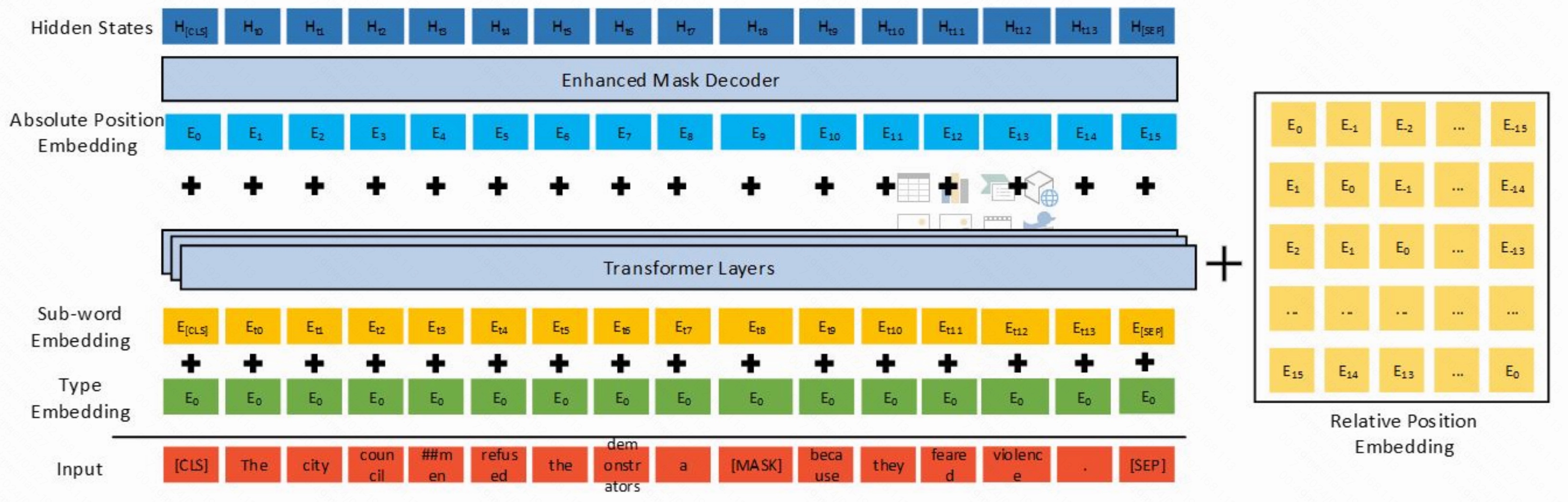}
    \caption{The architecture of DeBERTa \cite{he2021microsoft}}
    \label{fig:deberta}
\end{figure}

DeBERTa utilizes several layers of self-attention mechanisms and feed-forward neural networks in its architecture. The model architecture involves encoding the input tokens into contextualized representations and decoding them to generate predictions, as shown in Fig \ref{fig:deberta}. It utilizes disentangled attention mechanisms to capture token dependencies and learn contextual representations effectively.

Let $X = (x_1, x_2, ..., x_n)$ be the input token sequence. DeBERTa computes attention scores for each pair of tokens using query, key, and value matrices. The scores are then used to compute weighted sums of the values, producing context-aware representations for each token.

The self-attention mechanism in DeBERTa is defined as follows:

\begin{equation}
\text{Attention}(Q, K, V) = \text{softmax} \left( \frac{QK^T}{\sqrt{d}} \right) V
\end{equation}

Where:
\begin{itemize}
  \item $Q$, $K$, and $V$ are query, key, and value matrices respectively.
  \item $d$ is the dimensionality of the key vectors.
\end{itemize}

After self-attention layers, DeBERTa uses feed-forward networks to process the contextual representations. This typically involves element-wise operations followed by nonlinear activation functions.

Throughout the model architecture, layer normalization and residual connections are employed to stabilize training and facilitate information flow across layers.

During decoding or inference, DeBERTa generates predictions by applying additional layers to the contextual representations and applying appropriate output transformations based on the task at hand, such as classification or sequence generation.

\subsection{Model Ensemble}
The ensemble methodology employed in this study encompasses two pivotal phases, each meticulously designed to enhance the predictive performance of the overarching framework.

\subsubsection{TF-IDF Feature Extraction and Multi-Model Ensemble}

The initial phase of our ensemble architecture leverages TF-IDF feature extraction in conjunction with a sophisticated ensemble of classifiers to process the data and derive predictive outcomes. Upon TF-IDF feature extraction, model training proceeds with a meticulously chosen ensemble of classifiers, which includes prominent algorithms such as CatBoost and LightGBM. This ensemble learning framework is facilitated by the VotingClassifier, an advanced meta-learner orchestrating the weighted aggregation of predictions from individual classifiers. By amalgamating the outputs of diverse classifiers, this ensemble methodology not only mitigates the biases inherent in individual models but also significantly enhances the overall predictive accuracy.

Noteworthy is our empirical optimization effort, involving adjustments such as increasing the iteration count for classifiers like CatBoost and LightGBM, complemented by careful weight allocation. These refinements contribute synergistically to augment the predictive efficacy of the ensemble.

\subsubsection{Deberta-v3-large Model Training}

The methodology involves training twelve Deberta-v3-large models on diverse datasets and integrating them through ensemble techniques.

 After the initial training phase, optimization efforts target additional datasets, namely Pile and slimpajama. These datasets undergo rigorous filtering based on various criteria, including text length and presence of code or mathematical symbols.
 
Approximately 35 open-source models with diverse parameter combinations are employed for optimization on the Pile and slimpajama datasets. This diverse model selection enhances the robustness of the ensemble by capturing a wide range of textual nuances.

Further refinement involves fine-tuning five models on the selected 11K dataset. Although resulting in slightly higher public scores, these fine-tuned models exhibit marginally lower scores, indicating a trade-off between performance and generalization.

In the final step, we combine the results from these two parts with the ensemble outcome to bolster the model's robustness.

\subsection{Evaluation Metric: ROC-AUC}

The ROC-AUC score, essential for assessing binary classifiers, reflects their skill in differentiating between classes at various thresholds. This metric is derived from the model's True Positive Rate (TPR) and False Positive Rate (FPR), with higher values signifying superior performance.

\begin{equation}
\text{TPR} = \frac{\text{True Positives}}{\text{True Positives} + \text{False Negatives}}
\end{equation}

\begin{equation}
\text{FPR} = \frac{\text{False Positives}}{\text{False Positives} + \text{True Negatives}}
\end{equation}

The ROC curve is generated by plotting the true positive rate (TPR) versus the false positive rate (FPR) at different thresholds. The area under this curve is known as ROC-AUC:

\begin{equation}
\text{ROC-AUC} = \int_{0}^{1} \text{TPR}(f) \, d\text{FPR}(f)
\end{equation}

The ROC-AUC metric varies between 0 and 1, where 1 signifies flawless classification, and 0.5 represents a guesswork level of accuracy. It provides a comprehensive assessment of model performance across different threshold settings, particularly useful when dealing with class imbalances or when different operating points are desired.

\section{Experimental Results}
As shown in the table \ref{tab:results}, the ensemble model incorporating TF-IDF with Bayesian classifiers, Stochastic Gradient Boosting, CatBoost, and LightGBM, along with 12 instances of Deberta-v3-large, achieves the highest ROC-AUC score of 0.975022. This demonstrates its enhanced ability to differentiate between text created by humans and AI, surpassing alternative setups.

These results highlight the effectiveness of our proposed approach, particularly the ensemble of advanced models, in detecting AI-generated text with high accuracy.

\begin{table}[h]
    \caption{Experimental Results}
    \centering
    \begin{tabular}{|c|c|}
        \hline
        \textbf{Model} & \textbf{ROC-AUC} \\
        \hline
        TF-IDF + TransformerCNN + Roberta & 0.898283 \\
                \hline
        Bayesian/sgd/svc/lgb ensemble Classifier & 0.916748 \\
                \hline
        TF-IDF + Riadge/LinearSVR + DistilRoberta & 0.957233 \\
                \hline
        TF-IDF + Baysian/sgb/cat/lgb + 12*Deber-v3-large & 0.975022 \\
        \hline
    \end{tabular}
    \label{tab:results}
\end{table}

\section{Conclusion}
The proliferation of AI-generated text presents significant challenges in maintaining trust and authenticity in online communication. In response, we proposed a novel hybrid approach that combines traditional TF-IDF techniques with advanced machine learning models, including Bayesian classifiers, Stochastic Gradient Descent (SGD), Categorical Gradient Boosting (CatBoost), and Deberta-v3-large models.Our experimental results validate our method's ability to differentiate human from AI-generated text, attaining an impressive ROC-AUC score of 0.975022. Our ensemble methodology leverages the strengths of both traditional feature extraction methods and state-of-the-art deep learning models, contributing to the advancement of AI-generated text detection techniques. By mitigating the risks associated with AI-generated content, our research lays the foundation for fostering trust and authenticity in digital communication platforms.

 \bibliographystyle{IEEEtran}
    \bibliography{references}

\begin{thebibliography}{10}
\providecommand{\url}[1]{#1}
\csname url@samestyle\endcsname
\providecommand{\newblock}{\relax}
\providecommand{\bibinfo}[2]{#2}
\providecommand{\BIBentrySTDinterwordspacing}{\spaceskip=0pt\relax}
\providecommand{\BIBentryALTinterwordstretchfactor}{4}
\providecommand{\BIBentryALTinterwordspacing}{\spaceskip=\fontdimen2\font plus
\BIBentryALTinterwordstretchfactor\fontdimen3\font minus \fontdimen4\font\relax}
\providecommand{\BIBforeignlanguage}[2]{{%
\expandafter\ifx\csname l@#1\endcsname\relax
\typeout{** WARNING: IEEEtran.bst: No hyphenation pattern has been}%
\typeout{** loaded for the language `#1'. Using the pattern for}%
\typeout{** the default language instead.}%
\else
\language=\csname l@#1\endcsname
\fi
#2}}
\providecommand{\BIBdecl}{\relax}
\BIBdecl

\bibitem{bao2023fast}
G.~Bao, Y.~Zhao, Z.~Teng, L.~Yang, and Y.~Zhang, ``Fast-detectgpt: Efficient zero-shot detection of machine-generated text via conditional probability curvature,'' \emph{arXiv preprint arXiv:2310.05130}, 2023.

\bibitem{song2022comparative}
B.~Song and Y.~Zhao, ``A comparative research of innovative comparators,'' in \emph{Journal of Physics: Conference Series}, vol. 2221, no.~1.\hskip 1em plus 0.5em minus 0.4em\relax IOP Publishing, 2022, p. 012021.

\bibitem{dong2024prediction}
X.~Dong, B.~Dang, H.~Zang, S.~Li, and D.~Ma, ``The prediction trend of enterprise financial risk based on machine learning arima model,'' \emph{Journal of Theory and Practice of Engineering Science}, vol.~4, no.~01, pp. 65--71, 2024.

\bibitem{chen2024taskclip}
H.~Chen, W.~Huang, Y.~Ni, S.~Yun, F.~Wen, H.~Latapie, and M.~Imani, ``Taskclip: Extend large vision-language model for task oriented object detection,'' \emph{arXiv preprint arXiv:2403.08108}, 2024.

\bibitem{huang2023approximating}
C.~Huang, Y.~Xie, Z.~Jiang, J.~Lin, and M.~Li, ``Approximating human-like few-shot learning with gpt-based compression,'' \emph{arXiv preprint arXiv:2308.06942}, 2023.

\bibitem{lu2024m2fnet}
Y.~Lu, Y.~Huang, S.~Sun, T.~Zhang, X.~Zhang, S.~Fei, and V.~Chen, ``M2fnet: Multi-modal forest monitoring network on large-scale virtual dataset,'' \emph{arXiv preprint arXiv:2402.04534}, 2024.

\bibitem{xu2023automated}
J.~Xu, Y.~Jiang, B.~Yuan, S.~Li, and T.~Song, ``Automated scoring of clinical patient notes using advanced nlp and pseudo labeling,'' in \emph{2023 5th International Conference on Artificial Intelligence and Computer Applications (ICAICA)}.\hskip 1em plus 0.5em minus 0.4em\relax IEEE, 2023, pp. 384--388.

\bibitem{akram2023empirical}
A.~Akram, ``An empirical study of ai generated text detection tools,'' \emph{arXiv preprint arXiv:2310.01423}, 2023.

\bibitem{kaiyu2014new}
W.~Kaiyu, W.~Yumei, and Z.~Lin, ``A new three-layer qoe modeling method for http video streaming over wireless networks,'' in \emph{2014 4th IEEE International Conference on Network Infrastructure and Digital Content}.\hskip 1em plus 0.5em minus 0.4em\relax IEEE, 2014, pp. 56--60.

\bibitem{hong2024wildfake}
Y.~Hong and J.~Zhang, ``Wildfake: A large-scale challenging dataset for ai-generated images detection,'' \emph{arXiv preprint arXiv:2402.11843}, 2024.

\bibitem{han2024chain}
G.~Han, W.~Liu, X.~Huang, and B.~Borsari, ``Chain-of-interaction: Enhancing large language models for psychiatric behavior understanding by dyadic contexts,'' \emph{arXiv preprint arXiv:2403.13786}, 2024.

\bibitem{diao2023deep}
S.~Diao, W.~Luo, J.~Hou, R.~Lambo, H.~A. Al-Kuhali, H.~Zhao, Y.~Tian, Y.~Xie, N.~Zaki, and W.~Qin, ``Deep multi-magnification similarity learning for histopathological image classification,'' \emph{IEEE Journal of Biomedical and Health Informatics}, vol.~27, no.~3, pp. 1535--1545, 2023.

\bibitem{wang2012new}
C.~Wang, L.~Sun, J.~Wei, and X.~Mo, ``A new trojan horse detection method based on negative selection algorithm,'' in \emph{Proceedings of 2012 IEEE International Conference on Oxide Materials for Electronic Engineering (OMEE)}, 2012, pp. 367--369.

\bibitem{wang2010identification}
C.~Wang, H.~Yang, Y.~Chen, L.~Sun, Y.~Zhou, and H.~Wang, ``Identification of image-spam based on sift image matching algorithm,'' \emph{JOURNAL OF INFORMATION \&COMPUTATIONAL SCIENCE}, vol.~7, no.~14, pp. 3153--3160, 2010.

\bibitem{lim2024effect}
S.~Lim and R.~Schm{\"a}lzle, ``The effect of source disclosure on evaluation of ai-generated messages: A two-part study,'' \emph{Computers in Human Behavior: Artificial Humans}, p. 100058, 2024.

\bibitem{ma2023implementation}
D.~Ma, B.~Dang, S.~Li, H.~Zang, and X.~Dong, ``Implementation of computer vision technology based on artificial intelligence for medical image analysis,'' \emph{International Journal of Computer Science and Information Technology}, vol.~1, no.~1, pp. 69--76, 2023.

\bibitem{hu2024radar}
X.~Hu, P.-Y. Chen, and T.-Y. Ho, ``Radar: Robust ai-text detection via adversarial learning,'' \emph{Advances in Neural Information Processing Systems}, vol.~36, 2024.

\bibitem{zhang2023trep}
Z.~Zhang, R.~Tian, and Z.~Ding, ``Trep: Transformer-based evidential prediction for pedestrian intention with uncertainty,'' in \emph{Proceedings of the AAAI Conference on Artificial Intelligence}, vol.~37, 2023.

\bibitem{walters2023effectiveness}
W.~H. Walters, ``The effectiveness of software designed to detect ai-generated writing: A comparison of 16 ai text detectors,'' \emph{Open Information Science}, vol.~7, no.~1, p. 20220158, 2023.

\bibitem{chaka2023detecting}
C.~Chaka, ``Detecting ai content in responses generated by chatgpt, youchat, and chatsonic: The case of five ai content detection tools,'' \emph{Journal of Applied Learning and Teaching}, vol.~6, no.~2, 2023.

\bibitem{cingillioglu2023detecting}
I.~Cingillioglu, ``Detecting ai-generated essays: the chatgpt challenge,'' \emph{The International Journal of Information and Learning Technology}, vol.~40, no.~3, pp. 259--268, 2023.

\bibitem{mchugh2023defensive}
J.~McHugh, ``Defensive ai: Experimental study,'' Ph.D. dissertation, Marymount University, 2023.

\bibitem{sun2024securing}
L.~Sun, ``Securing supply chains in open source ecosystems: Methodologies for determining version numbers of components without package management files,'' \emph{Journal of Computing and Electronic Information Management}, vol.~12, no.~1, pp. 32--36, 2024.

\bibitem{varma2021systematic}
R.~Varma, Y.~Verma, P.~Vijayvargiya, and P.~P. Churi, ``A systematic survey on deep learning and machine learning approaches of fake news detection in the pre-and post-covid-19 pandemic,'' \emph{International Journal of Intelligent Computing and Cybernetics}, vol.~14, no.~4, pp. 617--646, 2021.

\bibitem{dang2024enhancing}
B.~Dang, D.~Ma, S.~Li, X.~Dong, H.~Zang, and R.~Ding, ``Enhancing kitchen independence: Deep learning-based object detection for visually impaired assistance,'' \emph{Academic Journal of Science and Technology}, vol.~9, no.~2, pp. 180--184, 2024.

\bibitem{wang2016quality}
Y.~Wang, M.~Sun, K.~Wang, and L.~Zhang, ``Quality of experience estimation with layered mapping for hypertext transfer protocol video streaming over wireless networks,'' \emph{International Journal of Communication Systems}, vol.~29, no.~14, pp. 2084--2099, 2016.

\bibitem{zang2024evaluating}
H.~Zang, S.~Li, X.~Dong, D.~Ma, and B.~Dang, ``Evaluating the social impact of ai in manufacturing: A methodological framework for ethical production,'' \emph{Academic Journal of Sociology and Management}, vol.~2, no.~1, pp. 21--25, 2024.

\bibitem{qaiser2018text}
S.~Qaiser and R.~Ali, ``Text mining: use of tf-idf to examine the relevance of words to documents,'' \emph{International Journal of Computer Applications}, vol. 181, no.~1, pp. 25--29, 2018.

\bibitem{john2013estimating}
G.~H. John and P.~Langley, ``Estimating continuous distributions in bayesian classifiers,'' \emph{arXiv preprint arXiv:1302.4964}, 2013.

\bibitem{li2021feature}
Z.-s. LI, X.~YAO, Z.-g. LIU, and J.-c. ZHANG, ``Feature selection algorithm based on lightgbm,'' \emph{Journal of Northeastern University (Natural Science)}, vol.~42, no.~12, p. 1688, 2021.

\bibitem{hancock2020catboost}
J.~T. Hancock and T.~M. Khoshgoftaar, ``Catboost for big data: an interdisciplinary review,'' \emph{Journal of big data}, vol.~7, no.~1, pp. 1--45, 2020.

\bibitem{he2021microsoft}
P.~He, X.~Liu, J.~Gao, and W.~Chen, ``Microsoft deberta surpasses human performance on the superglue benchmark,'' \emph{Microsoft, Redmond. Accessed: Nov}, vol.~18, 2021.

\end{thebibliography}

\end{document}